\documentclass[conference]{IEEEtran}
\IEEEoverridecommandlockouts
\usepackage{cite}
\usepackage{url}
\usepackage{makecell}

\usepackage{amsmath,amssymb,amsfonts}
\usepackage{algorithmic}
\usepackage{graphicx}
\usepackage{textcomp}
\usepackage{xcolor}
\usepackage{booktabs}
\usepackage{amsmath}
\usepackage{amssymb}
\usepackage{enumitem}
\usepackage{amsthm}
\usepackage{multirow}
\def\BibTeX{{\rm B\kern-.05em{\sc i\kern-.025em b}\kern-.08em
    T\kern-.1667em\lower.7ex\hbox{E}\kern-.125emX}}
  
\begin{document}

\title{Instance-Aware Parameter Configuration in Bilevel Late Acceptance Hill Climbing for the Electric Capacitated Vehicle Routing Problem \\
\thanks{The corresponding data and source code are available at
\protect\url{https://github.com/KingQino/Instance-aware-algorithm-configuration}.}
}

% \title{Instance-Aware Parameter Configuration in Bilevel Late Acceptance Hill Climbing for the Electric Capacitated Vehicle Routing Problem \\
% \thanks{The corresponding data and source code are omitted for blind review.}
% }

\author{
\IEEEauthorblockN{Yinghao Qin, Xinwei Wang, Jun Chen$^\star$}
\IEEEauthorblockA{\textit{Centre for Intelligent Transport} \\
\textit{Queen Mary, University of London}\\
London, UK \\
\{y.qin, xinwei.wang, jun.chen\}@qmul.ac.uk}
\thanks{$^\star$Corresponding author}
\and
\IEEEauthorblockN{Mosab Bazargani}
\IEEEauthorblockA{\textit{School of Computer Science and Engineering} \\
\textit{Bangor University}\\
Bangor, UK \\
m.bazargani@bangor.ac.uk}
}

% \author{
% \IEEEauthorblockN{Anonymous Authors}
% \IEEEauthorblockA{Anonymous Affiliation}
% }

\maketitle

\begin{abstract}
Algorithm performance in combinatorial optimisation is highly sensitive to parameter settings, while a single globally tuned configuration often fails to exploit the heterogeneity of problem instances. This limitation is particularly evident in the Electric Capacitated Vehicle Routing Problem, where instances differ substantially in structure, demand patterns, and energy constraints. This paper investigates instance-aware parameter configuration for Bilevel Late Acceptance Hill Climbing, a state-of-the-art metaheuristic for the Electric Capacitated Vehicle Routing Problem. An offline tuning procedure is used to obtain instance-specific parameter labels, which are then mapped from instance features via a regression model to enable parameter prediction for unseen instances prior to execution. Experimental results on the IEEE WCCI 2020 benchmark and its extensions show that the proposed approach achieves an average objective value reduction of $0.28\%$ across eight held-out test instances relative to a globally tuned configuration. This corresponds to a significant cost reduction in multimillion-dollar transportation operations.
\end{abstract}


\section{Introduction}

When designing algorithms for combinatorial optimisation problems, a fundamental challenge lies in determining appropriate parameter settings~\cite{hutter2005parameter, birattari2009tuning, kadioglu2010isac, lopez2020automated}. In practice, algorithm performance is often highly sensitive to these configurations. More importantly, different problem instances typically exhibit heterogeneous structural properties, which in turn induce distinct search landscapes. Consequently, a parameter configuration that performs well on one instance may lead to unsatisfactory performance on another.

To achieve consistently good performance across diverse problem instances, two representative research directions can be identified. One direction focuses on \emph{adaptive mechanisms}~\cite{rodriguez2024new, lin2025q, wang2024deep, qin2024confidence}, in which algorithm parameters or behaviours are adjusted online in response to the evolving search dynamics. While such mechanisms are effective in reacting to feedback observed during the run, they primarily rely on online information.

% To achieve consistently good performance across diverse problem instances, two representative research directions can be identified. One direction focuses on \emph{adaptive mechanisms}~\cite{rodriguez2024new, lin2025q, wang2024deep, qin2024confidence}, in which algorithm parameters or behaviours are adjusted online in response to the evolving search dynamics. For example, when the search becomes trapped in a local optimum, acceptance criteria for new candidate solutions may be relaxed or perturbation strength intensified to promote exploration. While such mechanisms are effective in reacting to feedback observed during the run, they primarily rely on online information.

An alternative direction is \emph{instance-aware parameter configuration (IAPC)}~\cite{hutter2005parameter, kadioglu2010isac, visweswaran2010learning}, which aims to determine suitable parameter settings \emph{a priori} based on the intrinsic characteristics of a given problem instance. This paradigm assumes that a mapping exists between instance-level features and algorithm parameters, allowing the configuration to be specialised before the search starts.

Bilevel Late Acceptance Hill Climbing (b-LAHC) is a metaheuristic developed for the Electric Capacitated Vehicle Routing Problem (E-CVRP)~\cite{qin2024confidence}. Under a fixed evaluation budget, it has demonstrated strong performance on the IEEE WCCI 2020 EVRP benchmark~\cite{mavrovouniotis2020techreport}, attaining highly competitive solutions on seven small-scale instances and improving $9$ out of $10$ best-known results on large-scale benchmarks, with an average improvement of $1.23\%$. In its current form, b-LAHC involves five algorithmic parameters, which were configured globally and fixed across all benchmark instances in previous studies.

Although such a global configuration provides a strong and reliable baseline, it implicitly assumes that a single parameter setting can perform well across instances with diverse structural characteristics. However, E-CVRP instances exhibit pronounced heterogeneity in terms of network topology, customer demand distributions, and energy-related constraints. As a result, this assumption may limit the full potential of b-LAHC on certain categories of instances.

Motivated by this observation, this paper investigates instance-aware algorithm configuration for b-LAHC, aiming to tailor its parameter settings to the characteristics of individual E-CVRP instances. By exploiting instance-level information to guide parameter selection, the proposed approach seeks not only to further improve algorithm performance but also to provide deeper insights into the interaction between b-LAHC’s algorithmic behaviour and problem instance characteristics.

The main contributions of this paper are summarised as follows:
\begin{itemize}[leftmargin=1.2em]
   \item An instance-aware configuration framework for b-LAHC is proposed, in which a regression-based model learns the mapping between E-CVRP instance characteristics and algorithm parameter settings, achieving an average improvement of $0.28\%$ over a global configuration on the test set. This is closely aligned with what Bengio et al. (2021) \cite{BENGIO2021405} suggest regarding the use of machine learning to configure algorithm behaviour (``tuning'') in operational research.
  \item An interpretable instance-level analysis is provided to elucidate how structural and demand-related characteristics of E-CVRP instances influence effective b-LAHC parameter choices, offering practical insights into the interaction between instance heterogeneity and algorithm behaviour.
\end{itemize}

The remainder of the paper is organised as follows. Section~\ref{related-work} reviews the related literature. Section~\ref{background} introduces the E-CVRP and the b-LAHC algorithm. Section~\ref{instance-aware-parameter-prediction-framework} presents the proposed instance-aware parameter prediction framework. Section~\ref{experiments} reports the experimental study and discusses the performance of b-LAHC under different parameter configurations. Finally, Section~\ref{conclusion} concludes the paper and outlines directions for future research.

\section{Related Work}
\label{related-work}

Hyperparameter configuration plays a critical role in determining the performance of algorithms for combinatorial optimisation problems. Most existing state-of-the-art approaches adopt a \emph{global} parameter configuration~\cite{qin2024confidence, jia2021bilevel, jia2022confidence, wang2023dual, chen2024efficient, woller2025variable}, where a single set of parameters is tuned offline and then uniformly applied to all problem instances. While this strategy provides a robust and convenient baseline, it inherently ignores the structural and distributional heterogeneity among instances, a limitation that is consistent with the implications of the ``no free lunch'' theorem. To address the limitations of global parameter configuration, two main research directions have been explored.

The first direction focusses on \emph{adaptive mechanisms}, in which algorithmic behaviours or control parameters are adjusted online based on feedback from the ongoing search process. In the context of the E-CVRP, various adaptive strategies have been proposed, including self-adaptive acceptance criteria~\cite{rodriguez2024new}, reinforcement-learning-based operator selection~\cite{lin2025q, wang2024deep}, and adaptive population control mechanisms in population-based algorithms~\cite{qin2024confidence}.

% In the context of the E-CVRP, several adaptive strategies have been proposed. Rodriguez et al.~\cite{rodriguez2024new} introduced a self-adaptive simulated annealing approach that relaxes the acceptance criterion when the search stagnates and progressively tightens it otherwise. Hyper-heuristic frameworks, often implemented using reinforcement learning, have also been widely adopted to enable dynamic operator selection. For example, Lin et al.~\cite{lin2025q} employed Q-learning to adaptively select low-level heuristics during the search, while Wang et al.~\cite{wang2024deep} formulated operator and charging strategy selection as a Markov Decision Process and applied deep reinforcement learning to guide destroy and repair decisions. Adaptive ideas have further been incorporated into population-based algorithms. Qin et al.~\cite{qin2024confidence} proposed an adaptive selection mechanism that dynamically controls mating patterns among different groups of individuals to regulate convergence behaviour, biasing reproduction towards high-quality individuals in the early stages and gradually increasing diversity through interactions with average individuals and randomly generated solutions as the search progresses.

An alternative research direction is \emph{instance-aware algorithm configuration} (IAAC), which aims to determine suitable algorithmic choices for each problem instance \emph{before} the search begins. These choices may involve selecting algorithmic components, control strategies, or parameter values, depending on the scope of the configuration problem. Early work by Hutter et al.~\cite{hutter2005parameter} formally defined the per-instance algorithm configuration problem, extending the classical algorithm selection paradigm~\cite{rice1976algorithm} from selecting among algorithms to selecting configurations of a single algorithm. Subsequent studies have adopted a \emph{machine learning perspective}, demonstrating that instance-aware configuration can outperform globally optimised configurations by exploiting instance-level heterogeneity~\cite{visweswaran2010learning}.

% Instance-aware algorithm configuration (IAPC) has been extensively studied as a principled approach to exploit instance heterogeneity. Early work by Hutter et al.~\cite{hutter2005parameter} formally defined the per-instance algorithm configuration problem and proposed a performance-prediction-based framework, extending the classical algorithm selection paradigm~\cite{rice1976algorithm} from selecting among algorithms to selecting configurations of a single algorithm. Subsequent studies have adopted a machine learning perspective, demonstrating that instance-specific predictive models can outperform globally optimised models by tailoring decisions to individual instances~\cite{visweswaran2010learning}.

The effectiveness of IAAC has been demonstrated across several combinatorial optimisation domains. Ansótegui et al.~\cite{ansotegui2016maxsat} showed that instance-specific configuration can significantly improve solvers for the Maximum Satisfiability (MaxSAT) problem across heterogeneous instance families. More recently, Song et al.~\cite{song2023instance} proposed an instance-aware configuration framework based on unsupervised deep graph clustering, yielding substantial performance gains for mixed-integer programming (MIP) solvers by automatically capturing structural differences among instances. Related ideas have also been explored in dynamic settings, where instance-specific configurations or solver portfolios are evolved over time to cope with changing instance distributions~\cite{malitsky2013evolving}.

Despite its success in domains such as SAT, MaxSAT, and MIP, instance-aware algorithm configuration has received little attention in the vehicle routing problem community. This gap motivates the present study, which investigates instance-aware \emph{parameter} configuration as a specific instantiation of IAAC for b-LAHC in the E-CVRP.

\begin{figure}
  \centering
  \includegraphics[width=0.9\linewidth]{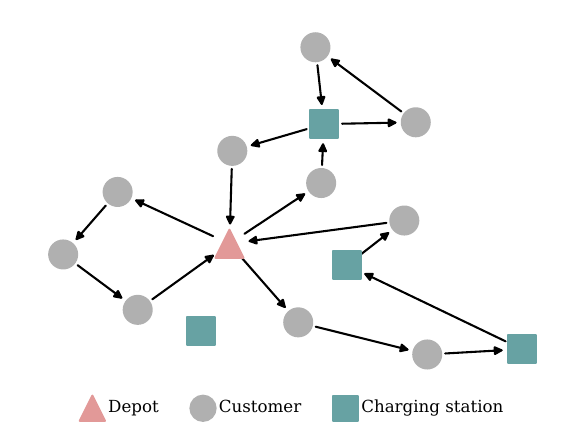}
  \caption{Illustrative solution to the E-CVRP.}
  \label{fig:example-solution}
\end{figure}

\section{Background}\label{background}

\subsection{Electric Capacitated Vehicle Routing Problem}
\label{subsec:E-CVRP}

The E-CVRP concerns the routing of a homogeneous fleet of electric vehicles (EVs) to serve a set of geographically distributed customers. The objective is to minimise the total travelled distance while satisfying customer demands and respecting both vehicle load capacity and battery energy constraints. Each vehicle departs from a central depot, visits a subset of customers, and returns to the depot upon completion of its route.

Unlike the classical Capacitated Vehicle Routing Problem (CVRP), the E-CVRP explicitly accounts for energy consumption and recharging decisions. Vehicles may visit charging stations to replenish their batteries, potentially multiple times along a route, in order to remain feasible with respect to energy constraints. As a result, route feasibility depends not only on customer sequencing but also on the placement and ordering of charging station visits.

Fig.~\ref{fig:example-solution} illustrates an example of a feasible E-CVRP solution comprising three routes, highlighting different charging patterns, ranging from routes without recharging to routes involving multiple and consecutive charging station visits.

A complete problem definition and mathematical formulation of the E-CVRP can be found in~\cite{qin2024confidence}.

\begin{table}[!t]
\centering
\caption{Key b-LAHC parameters for instance-aware prediction}
\label{tab:parameters}
\renewcommand{\arraystretch}{0.8}
\begin{tabular}{llc}
\toprule
\textbf{Parameter} & \textbf{Description} & \textbf{Global value} \\
\midrule
$L_h$ & History length                                & 5723 \\
$\eta_{\max}$  & Maximum attempts                      & 60   \\
\bottomrule
\end{tabular}
\end{table}

\begin{figure*}[!t]
	\centering
	\includegraphics[width=0.8\linewidth]{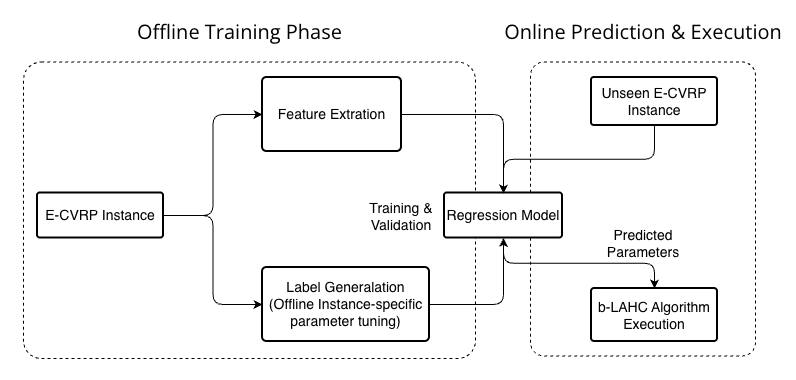}
	\caption{Overview of the proposed instance-aware parameter prediction framework for b-LAHC.}
	\label{fig:framework}
\end{figure*}

\subsection{Bilevel Late Acceptance Hill Climbing}
\label{subsec:b-LAHC}

The b-LAHC is a metaheuristic designed for solving the E-CVRP. It is built upon a path-based bilevel optimisation model, in which upper-level decisions determine vehicle routes, while lower-level decisions optimise charging strategies along each route, reflecting the intrinsic coupling between routing and recharging decisions.

The algorithm is derived from Late Acceptance Hill Climbing (LAHC)~\cite{burke2017late}, a single-point metaheuristic widely applied in scheduling and combinatorial optimisation~\cite{fonseca2016late, bolaji2018late}. The core idea of LAHC is the \emph{late acceptance} criterion, under which a candidate solution is accepted if it improves upon the current solution or is better than a solution encountered several iterations earlier.

The b-LAHC algorithm is controlled by five parameters, among which the history length $L_h$ and the maximum number of attempts per heuristic operator $\eta_{\max}$ have been identified as the most influential in previous sensitivity analyses. Table~\ref{tab:parameters} reports the globally tuned values of these two parameters obtained on the IEEE WCCI 2020 benchmark. This global configuration provides robust performance across heterogeneous instances but does not explicitly account for instance-specific characteristics.

The parameter $L_h$ determines the length of the historical list used in the late acceptance criterion. Larger values of $L_h$ promote \emph{exploration} by allowing the acceptance of worse solutions over a longer horizon, whereas smaller values favour \emph{exploitation}, leading to faster convergence at the risk of premature stagnation. The parameter $\eta_{\max}$ controls the intensity of neighbourhood exploration by limiting the number of candidate moves evaluated for each operator before switching to the next one. Larger values increase the likelihood of finding an acceptable move at the cost of higher evaluation effort per operator, whereas smaller values lead to more frequent operator switching or algorithm restarts under a fixed evaluation budget. Taken together, these two parameters jointly govern the exploration–exploitation trade-off and the effective use of the evaluation budget, making them natural targets for instance-aware configuration.

\begin{table*}[!t]
\centering
\caption{Summary of instance features used for instance-aware parameter prediction}
\label{tab:instance-features}
\renewcommand{\arraystretch}{1.0}
\begin{tabular}{lll}
\toprule
\textbf{Category} & \textbf{Feature} & \textbf{Description} \\
\midrule
\multirow{7}{*}{Basic information (7)}
 & num\_customers & Number of customer nodes to be served \\
 & num\_stations & Number of available charging stations \\
 & num\_depot & Number of depot nodes \\
 & num\_vehicles & Number of available vehicles in the fleet \\
 & vehicle\_capacity & Maximum load capacity of each vehicle \\
 & battery\_capacity & Maximum battery energy capacity of each vehicle \\
 & energy\_consumption & Energy consumption rate per unit distance \\
\midrule
\multirow{25}{*}{Graph-based features (25)}
 & depot\_in\_giant\_component & Indicates whether the depot belongs to the giant component of the kNN graph \\
 & is\_connected & Indicates whether the constructed kNN graph is fully connected \\
 & N\_giant / M\_giant & Number of nodes / edges in the giant component \\
 & deg\_mean / deg\_std & Mean and standard deviation of node degrees in the kNN graph \\
 & deg\_min / deg\_max & Minimum and maximum node degree in the kNN graph \\
 & deg\_gc\_mean / deg\_gc\_std & Degree statistics within the giant component \\
 & edge\_w\_gc\_mean / std / min / max & Edge weight statistics within the giant component \\
 & clust\_gc\_mean / clust\_gc\_std & Mean and dispersion of clustering coefficients in the giant component \\
 & mst\_weight & Total edge weight of the MST \\
 & mst\_weight\_per\_node & MST weight normalised by the number of nodes \\
 & mst\_deg\_mean / mst\_deg\_max & Degree statistics of nodes in the MST \\
 & avg\_shortest\_path\_w & Average weighted shortest path length in the giant component \\
 & diameter\_unweighted & Unweighted diameter of the giant component \\
 & depot\_betweenness\_w & Weighted betweenness centrality of the depot node \\
 & depot\_closeness\_w & Weighted closeness centrality of the depot node \\
 & degree\_assortativity & Degree assortativity coefficient of the giant component \\
 & lap\_eig\_min / max / mean / std & Statistics of normalised Laplacian eigenvalues \\
 & algebraic\_connectivity & Second smallest eigenvalue of the normalised Laplacian \\
 & pairdist\_mean / std / cv & Statistics of pairwise Euclidean distances between nodes \\
 & nn\_dist\_mean / std / cv & Statistics of nearest-neighbour distances \\
 & cust\_to\_station\_nn\_mean / std & Distance statistics from customers to their nearest charging stations \\
\midrule
\multirow{8}{*}{Demand-related features (8)}
 & demand\_mean / demand\_std / demand\_cv & Mean, dispersion, and coefficient of variation of customer demands \\
 & demand\_to\_capacity\_mean & Mean customer demand normalised by vehicle capacity \\
 & demand\_to\_capacity\_max & Maximum demand-to-capacity ratio \\
 & total\_demand\_to\_capacity & Total customer demand normalised by vehicle capacity \\
 & demand\_skewness & Skewness of the customer demand distribution \\
 & demand\_entropy & Entropy of the customer demand distribution \\
 & high\_demand\_ratio & Proportion of customers whose demand exceeds a half of vehicle capacity \\
 & demand\_weighted\_station\_dist & Demand-weighted distance from customers to nearest charging stations \\
\bottomrule
\end{tabular}
\end{table*}

\section{Instance-aware Parameter Prediction Framework}\label{instance-aware-parameter-prediction-framework}

\subsection{Overview of the Framework}

This section provides an overview of the proposed instance-aware parameter prediction framework for b-LAHC. The goal of the framework is to automatically infer effective algorithm parameter configurations for unseen E-CVRP instances based solely on their instance characteristics, without requiring any additional parameter tuning at execution time.

As illustrated in Fig.~\ref{fig:framework}, the framework consists of an offline training phase followed by an online inference phase. During offline training, each E-CVRP instance undergoes two parallel processes. First, instance-specific parameter tuning is conducted independently for each training instance to obtain near-optimal parameter configurations, which serve as supervisory labels. Second, a set of descriptive features, detailed in Table~\ref{tab:instance-features}, is extracted to characterise the structural and demand-related properties of each instance. Based on the resulting feature--label pairs, a regression model is trained to learn the mapping from instance characteristics to effective b-LAHC parameters.

Once trained, the regression model enables straightforward parameter configuration for new instances. For an unseen E-CVRP instance, the same feature extraction procedure is applied and an instance-specific parameter configuration is predicted and directly supplied to b-LAHC, which is executed under the same stopping criterion as the baseline configuration, without any further parameter adaptation.

Importantly, instance-specific parameter tuning is required only during the offline training stage, enabling efficient and scalable instance-aware parameter configuration when solving new E-CVRP instances.

\subsection{Offline Instance-specific Parameter Tuning}
\label{subsec:offline-tuning}

The objective of the offline tuning procedure is to obtain high-quality, near-optimal configurations of the b-LAHC parameters for individual E-CVRP instances. These configurations subsequently serve as reference labels for supervised parameter prediction.

A two-stage tuning strategy is employed using the automated configuration tool \textsc{irace}~\cite{lopez2016irace}. In the first stage, a coarse-grained exploration is conducted over a broad parameter domain, with $L_h \in [300, 20000]$ and $\eta_{\max} \in [1, 500]$. A fixed tuning budget of 3{,}500 configurations is allocated to this stage to obtain a coarse coverage of high-performing regions of the parameter space under a limited computational budget. The purpose of this coarse search is not to identify a single best configuration, but to locate high-performing regions of the parameter space.

Based on the top-performing configurations from the coarse stage, a refined search space is constructed using a gap-based multi-modality analysis. To detect potential multi-modality in $L_h$, the values are first log-transformed and sorted, and adjacent gaps are computed. If the maximum gap exceeds a predefined threshold $\tau$ ($\tau=0.8$), the parameter space is partitioned into multiple modes; otherwise, it is treated as unimodal. For each identified mode, the refinement interval of $L_h$ is defined using the empirical 10th and 90th percentiles, expanded by a 20\% buffer for robustness.

For $\eta_{\max}$, whose feasible range is comparatively small, multi-modality is not considered. Instead, a single refinement interval is constructed using the same quantile-based rule. For each mode identified by $L_h$, 8{,}000 parameter configurations are evaluated jointly over $(L_h,\eta_{\max})$ during the fine-grained tuning stage.

After fine-grained tuning, the top-$K$ configurations for each mode are aggregated using robust statistics. For each parameter, the median of the top-$K$ configurations is taken as the representative value, with the interquartile range (Q1--Q3) used to characterise the concentration of high-performing configurations. For instances exhibiting multiple modes, the mode achieving the best performance under b-LAHC is selected, and the resulting median configuration is used as the instance-specific parameter label for subsequent model training.

\subsection{Instance Feature Extraction}
\label{subsec:instance-features}

To support instance-aware parameter prediction, each E-CVRP instance is encoded using a set of descriptive features capturing its key structural and operational properties. The features are grouped into three categories: basic instance descriptors, graph-based structural features, and demand-related features. A complete list of features and their definitions is summarised in Table~\ref{tab:instance-features}.

Graph-based features are derived from a sparse $k$-nearest-neighbour (kNN) graph constructed from node coordinates, with $k=10$ and mutual connectivity enforced. Compared with the complete graph underlying the E-CVRP definition, this representation preserves local neighbourhood structure while avoiding degenerate graph statistics.

As the resulting kNN graph is not necessarily connected, structural features are computed on different graph substructures as appropriate. Local connectivity measures and degree-based statistics are extracted from the full kNN graph, while global properties requiring connectivity, including clustering coefficients, shortest-path-based measures, minimum spanning tree (MST) characteristics, centrality metrics, assortativity, and spectral features, are computed on the giant component. In addition, selected geometric distance statistics are derived directly from the Euclidean distance matrix.

Demand-related features characterise the distribution of customer demands and their relationship to vehicle capacity, including summary statistics, capacity tightness indicators, and distributional shape descriptors. Furthermore, a demand-weighted distance to the nearest charging station is also included to capture the interaction between demand patterns and charging infrastructure.

\subsection{Regression-based Parameter Prediction Model}
\label{subsec:regression-model}

Given the instance-specific parameter labels obtained through offline tuning and the corresponding instance features, a regression-based model is constructed to predict effective b-LAHC parameter settings for unseen E-CVRP instances \emph{prior} to execution, thus avoiding any online parameter adjustment.

Let $\mathbf{x}_i \in \mathbb{R}^d$ denote the feature vector extracted from instance $i$, and $\mathbf{y}_i \in \mathbb{R}^p$ the associated parameter configuration label. In this study, $p=2$, corresponding to the history length $L_h$ and the maximum attempt parameter $\eta_{\max}$. All features are standardised prior to training.

Ridge regression is adopted as the prediction model due to its robustness under limited training data and its ability to mitigate multicollinearity among correlated instance features. For the history length parameter, the predicted value is given by
\begin{equation}
\hat{L}_h = \beta_0 + \sum_{j=1}^{d} \beta_j \tilde{x}_j,
\end{equation}
where $\tilde{x}_j$ denotes the standardised instance features and $\beta_j$ are the learned regression coefficients. An analogous model is trained independently for $\eta_{\max}$. As the two parameters are modelled separately, no joint normalisation of the target variables is required; each regression model directly learns the mapping to the native scale of its corresponding parameter.

The model performance is assessed using $5$-fold cross-validation on the training instances, after which the trained regression model is used to predict parameter configurations for unseen instances and directly supplied to b-LAHC under an identical evaluation budget.

\section{Experiments}\label{experiments}

\begin{table}[!t]
\centering
\caption{Held-out test instances used for evaluating generalisation performance}
\label{tab:test-instances}
\renewcommand{\arraystretch}{0.85}
\begin{tabular}{ll}
\toprule
\textbf{Instance} & \textbf{Source set} \\
\midrule
E-n22-k4        & \cite{christofides1969algorithm} \\
E-n112-k8-s11   & \cite{christofides1969algorithm} \\
M-n212-k16-s12  & \cite{christofides1981exact} \\
F-n140-k5-s5    & \cite{fisher1994optimal} \\
X-n221-k11-s7   & \cite{uchoa2017new} \\
X-n469-k26-s10  & \cite{uchoa2017new} \\
X-n698-k75-s13  & \cite{uchoa2017new} \\
X-n1006-k43-s5  & \cite{uchoa2017new} \\
\bottomrule
\end{tabular}
\end{table}

\subsection{Benchmark Suite and Data Splitting}

The experimental evaluation is conducted on a widely used benchmark suite for the E-CVRP. In this study, a total of $41$ benchmark instances are considered, comprising the $17$ original instances of the IEEE WCCI 2020 EVRP benchmark introduced in~\cite{mavrovouniotis2020techreport}, together with $24$ additional instances from its extended benchmark suite~\cite{mavrovouniotis2020benchmark}. These instances cover a broad spectrum of problem scales and structural characteristics, involving different numbers of customers and charging stations, heterogeneous demand distributions, and diverse spatial layouts.

The benchmark instances originate from four classical CVRP source sets~\cite{christofides1969algorithm, christofides1981exact, fisher1994optimal, uchoa2017new}, distinguished by their instance name prefixes \texttt{E}, \texttt{M}, \texttt{F}, and \texttt{X}, respectively. Specifically, the benchmark comprises $14$ instances from the \texttt{E} set, $4$ from the \texttt{M} set, $3$ from the \texttt{F} set, and $20$ from the \texttt{X} set.

To evaluate the generalisation capability of the proposed instance-aware parameter prediction framework, the full benchmark is partitioned into a training set and an independent held-out test set. The test set consists of $8$ representative instances, selected to ensure coverage of all four source sets. These instances are summarised in Table~\ref{tab:test-instances}.

The remaining $33$ instances form the training set and are used for model development via a $5$-fold cross-validation procedure. In each fold, instance-specific parameter labels are used to train the regression model, which is then evaluated on the held-out fold. After cross-validation, a final regression model is trained on the complete training set and applied to the independent test set to assess performance on previously unseen instances.

\subsection{Experimental Protocol}

The termination criterion is defined as a function of the problem size $pz$,
which is measured as the sum of the numbers of depots, customers, and charging stations in the instance.
The maximum number of evaluations is set to
\begin{equation}
\text{MaxEvals} = 25{,}000 \times pz,
\label{eq:max-evals}
\end{equation}
where a single evaluation of a candidate solution has a time complexity of $\mathcal{O}(pz^2)$.

All experiments involving the execution of b-LAHC were conducted on the Sulis high-performance computing (HPC) platform using dedicated AMD EPYC 7742 nodes (2.25\,GHz, 64-core processors). Each run was restricted to single-thread execution with a memory limit of 1\,GB. To ensure fair comparisons across instances and parameter configurations, all b-LAHC runs were performed under an identical evaluation budget defined by \emph{MaxEvals}. The experimental setup for b-LAHC was strictly controlled, including identical hardware, implementation language (C++), compiler (GCC~13.3.0), optimisation flag (\texttt{-O3}), and $10$ independent runs per instance.

Statistical significance was assessed using a paired Wilcoxon signed-rank test on instance-level mean objective values, with a significance level of $\alpha = 0.05$.

\begin{table}[!t]
\centering
\caption{Effectiveness of instance-specific parameter tuning for b-LAHC}
\label{tab:instance-specific-vs-global}
\renewcommand{\arraystretch}{0.85}
\begin{tabular}{lc}
\toprule
\textbf{Metric} & \textbf{Value} \\
\midrule
Total number of instances & 41 \\
Instances improved (mean objective) & $31 / 41$ \\
Instances unchanged & $6 / 41$ \\
Instances degraded & $4 / 41$ \\
Average improvement  & $-0.27\%$ \\
Best improvement & $-1.16\%$ \\
Worst degradation & $+0.27\%$ \\
\bottomrule
\end{tabular}
\end{table}

\begin{table}[!t]
\centering
\caption{Five-fold cross-validation results for parameter prediction (mean $\pm$ std)}
\label{tab:cv-results}
\renewcommand{\arraystretch}{0.85}
\begin{tabular}{lccc}
\toprule
\textbf{Target} & \textbf{MAE} & \textbf{RMSE} & \textbf{Spearman} \\
\midrule
$L_h$          & $0.4446 \pm 0.1173$ & $0.5872 \pm 0.1805$ & $0.634 \pm 0.221$ \\
$\eta_{\max}$  & $1.0622 \pm 0.6351$ & $1.6456 \pm 0.9933$ & $0.664 \pm 0.428$ \\
\bottomrule
\end{tabular}
\end{table}

\begin{table}[!t]
\centering
\caption{Predicted instance-specific parameters for the test set}
\label{tab:predicted-params-test}
\renewcommand{\arraystretch}{0.85}
\begin{tabular}{lcc}
\toprule
\textbf{Instance} & $\hat{L}_h$ & $\hat{\eta}_{\max}$ \\
\midrule
E-n22-k4        & 1994 & 43  \\
E-n112-k8-s11   & 1505 & 6   \\
F-n140-k5-s5    & 4400 & 9   \\
M-n212-k16-s12  & 2854 & 7   \\
X-n221-k11-s7   & 6095 & 283 \\
X-n469-k26-s10  & 5630 & 237 \\
X-n698-k75-s13  & 5549 & 486 \\
X-n1006-k43-s5  & 7018 & 503 \\
\bottomrule
\end{tabular}
\end{table}

\begin{table*}[!t]
\centering
\caption{Performance of b-LAHC with predicted parameters on the test set (gaps are relative to the global configuration).}
\label{tab:predicted-performance}
\renewcommand{\arraystretch}{1.0}
\begin{tabular}{llccccc}
\toprule
\multirow{2}{*}{\textbf{Instance}} & \multirow{2}{*}{\textbf{Metric}}
& \multicolumn{1}{c}{\textbf{Global configuration}}
& \multicolumn{2}{c}{\textbf{Predicted parameters}}
& \multicolumn{2}{c}{\textbf{Instance-specific tuning (oracle)}} \\
\cmidrule(lr){3-3} \cmidrule(lr){4-5} \cmidrule(lr){6-7}
&  & Obj & Obj & Gap & Obj & Gap \\
\midrule

\multirow{3}{*}{E-n22-k4}
& Best & 384.68 & 384.68 & -- & 384.68 & -- \\
& Mean & 385.10 & 385.18 & 0.02\% & 385.25 & 0.04\% \\
& Std  & 0.37   & 0.34   & -- & 0.30   & -- \\
\midrule

\multirow{3}{*}{E-n112-k8-s11}
& Best & 831.11 & 831.11 & -- & 831.11 & -- \\
& Mean & 835.48 & 833.83 & -0.20\% & 834.32 & -0.14\% \\
& Std  & 1.99   & 1.93   & -- & 1.74   & -- \\
\midrule

\multirow{3}{*}{F-n140-k5-s5}
& Best & 1190.03 & 1189.64 & -- & 1163.55 & -- \\
& Mean & 1196.41 & 1193.03 & -0.28\% & 1182.50 & -1.16\% \\
& Std  & 8.77    & 9.87    & -- & 16.76   & -- \\
\midrule

\multirow{3}{*}{M-n212-k16-s12}
& Best & 1310.63 & 1308.71 & -- & 1308.86 & -- \\
& Mean & 1317.90 & 1316.92 & -0.07\% & 1314.30 & -0.27\% \\
& Std  & 4.35    & 4.06    & -- & 3.26    & -- \\
\midrule

\multirow{3}{*}{X-n221-k11-s7}
& Best & 12070.33 & 12040.53 & -- & 12003.47 & -- \\
& Mean & 12133.67 & 12102.09 & -0.26\% & 12070.52 & -0.52\% \\
& Std  & 40.28    & 33.04    & -- & 38.10    & -- \\
\midrule

\multirow{3}{*}{X-n469-k26-s10}
& Best & 24504.60 & 24411.94 & -- & 24432.67 & -- \\
& Mean & 24571.46 & 24516.04 & -0.23\% & 24534.33 & -0.15\% \\
& Std  & 33.00    & 78.81    & -- & 61.36    & -- \\
\midrule

\multirow{3}{*}{X-n698-k75-s13}
& Best & 69015.41 & 68799.94 & -- & 68691.69 & -- \\
& Mean & 69181.12 & 68948.51 & -0.34\% & 68913.07 & -0.39\% \\
& Std  & 89.50    & 81.45    & -- & 153.97   & -- \\
\midrule

\multirow{3}{*}{X-n1006-k43-s5}
& Best & 74409.50 & 73723.30 & -- & 73795.07 & -- \\
& Mean & 74582.92 & 73981.14 & -0.81\% & 73946.40 & -0.85\% \\
& Std  & 144.79   & 148.75   & -- & 142.51   & -- \\

\bottomrule
\end{tabular}
\end{table*}

\subsection{Algorithm Performance with Instance-specific Parameters}

Table~\ref{tab:instance-specific-vs-global} summarises the comparative performance of the global configuration and instance-specific parameter tuning for b-LAHC. Overall, instance-specific tuning provides consistent performance gains across the benchmark set, confirming that tailoring parameters to individual instances yields a stronger performance than a single global setting.

The small number of marginal degradations observed can be explained by the stochastic nature of the \textsc{irace} tuning process and the finite configuration budget, which targets near-optimal rather than globally optimal parameters. Moreover, b-LAHC itself is stochastic, and minor variations in mean objective values may arise from run-to-run dynamics. Importantly, these degradations are negligible in magnitude and do not undermine the overall effectiveness of instance-specific parameter tuning.

\subsection{Predicted Parameters: Training, Inference, and Performance}

\subsubsection{Regression model training with 5-fold cross-validation}

The regression model is evaluated using $5$-fold cross-validation on the $33$ training instances. In each fold, the model is trained on four folds and evaluated on the remaining unseen fold, using instance features to predict the instance-specific parameter labels obtained from offline tuning.

Table~\ref{tab:cv-results} summarises the prediction performance for both target parameters, $L_h$ and $\eta_{\max}$, measured by mean absolute error (MAE), root mean squared error (RMSE), and Spearman’s rank correlation coefficient. While absolute prediction errors are non-negligible, particularly for $\eta_{\max}$, both parameters exhibit moderate to strong positive Spearman correlations, indicating that the model is able to preserve the relative ordering of instances in terms of their effective parameter values.

Overall, these results suggest that the regression model captures meaningful monotonic relationships between instance characteristics and tuned b-LAHC parameters. The practical effectiveness of the predicted parameters is subsequently evaluated through their impact on downstream algorithm performance.

\subsubsection{Predicted parameters on the test set}

After training the regression model on the full training set, instance-specific parameters are predicted for the held-out test instances using their extracted features. Table~\ref{tab:predicted-params-test} reports the predicted values of the two target parameters, $L_h$ and $\eta_{\max}$, for each test instance.

\subsubsection{b-LAHC performance with predicted parameters}

Table~\ref{tab:predicted-performance} reports the performance of b-LAHC when using parameters predicted by the proposed instance-aware regression model on the held-out test set. The predicted parameters improve the mean objective value on $7$ out of $8$ test instances. Averaged over all eight test instances, the objective value is reduced by $0.28\%$ relative to the global configuration. The only performance degradation is observed on instance \texttt{E-n22-k4}, where the mean objective value is higher than that of the global configuration by $0.02\%$.

A paired Wilcoxon signed-rank test conducted on the instance-level mean objective values indicates that the performance difference between the predicted parameters and the global configuration is statistically significant ($p=0.0156$).

Moreover, the performance achieved with predicted parameters is very close to that obtained with instance-specific (oracle) tuning, indicating that the regression model provides an effective approximation of the optimal parameter configurations. The minor degradation observed on one instance can be attributed to the stochastic nature of b-LAHC and the limited evaluation budget, which may lead to small performance fluctuations on rugged search landscapes.

\subsubsection{Feature importance analysis}

To interpret how instance characteristics influence the predicted parameter values, we analyse the coefficients of the final Ridge regression model trained on the full training set. Since all features are standardised prior to training, the magnitude of each coefficient reflects the relative importance of the corresponding feature, while its sign indicates the direction of influence.

For the history length parameter $L_h$, the most influential features mainly arise from the demand-related and graph-based categories, many of which are computed on the giant component of the kNN graph. In particular, demand entropy and the proportion of high-demand customers show strong positive associations with $L_h$, indicating that instances with more heterogeneous demand profiles tend to benefit from longer history lists. Several graph-based structural features, including clustering statistics, extreme edge weights, and spectral properties of the normalised Laplacian, also receive substantial weights, suggesting that both local connectivity patterns and global structural properties influence the effectiveness of late acceptance.

In contrast, basic size-related descriptors, such as the number of customers or vehicles, exhibit relatively small coefficients. This implies that structural and distributional characteristics are more informative than instance scale alone for selecting suitable values of $L_h$. Overall, the learned feature importances are consistent with the role of $L_h$ in regulating the exploration and exploitation balance in b-LAHC and provide evidence that meaningful relationships exist between instance characteristics and effective algorithm parameter settings.

\section{Conclusion and Future Work}
\label{conclusion}

This paper investigated instance-aware algorithm configuration for b-LAHC applied to the E-CVRP. While existing E-CVRP algorithms predominantly rely on globally tuned parameter settings, we showed that explicitly accounting for instance-level heterogeneity can lead to consistent performance improvements.

An instance-aware parameter prediction framework was proposed, in which high-quality instance-specific configurations obtained via offline automated tuning are learned through a regression model based on descriptive instance features. This enables effective parameter inference for unseen instances without any online tuning overhead.

Experimental results on the IEEE WCCI 2020 EVRP benchmark and its extended instances demonstrate that instance-specific tuning consistently outperforms a global configuration, and that parameters predicted by the proposed model achieve performance close to oracle configurations on held-out test instances. Feature analysis further reveals that demand heterogeneity and graph-based structural properties play a key role in determining effective late acceptance behaviour.

Future work will explore the integration of adaptive mechanisms into b-LAHC, by modelling neighbourhood operator selection as a Markov Decision Process, allowing search behaviour to be guided dynamically by the current search state.

% \section*{Acknowledgment}
% This section is omitted for blind review.

\section*{Acknowledgment}

The first author gratefully acknowledges support from the China Scholarship Council (Grant No. 202209110001). The authors thank the IT support team at QMUL. The manuscript preparation was assisted by an artificial intelligence tool (ChatGPT, version~5.2). The authors take full responsibility for the content.

\bibliographystyle{IEEEtran}
\bibliography{refs}

\end{document}